\documentclass[11pt]{article}

\input epsf %% at top of file

\usepackage{jair}
\usepackage{theapa}
\usepackage{graphicx}
\usepackage{colortab} % for shaded boxes
\input{pstricks.tex} %load the colors
\usepackage{lscape}
\usepackage[english,boxed,linesnumbered]{algorithm2e}

\usepackage{rotating}
\usepackage{caption}

\jairheading{29}{2007}{79-103}{09/06}{06/07}
\ShortHeadings{NP Animacy Identification for Anaphora Resolution}
{Orasan \& Evans} \firstpageno{79}

\begin{document}

\author{\name Constantin Or\u{a}san \email C.Orasan@wlv.ac.uk \\
        \name Richard Evans \email R.J.Evans@wlv.ac.uk \\
       \addr Research Group in Computational Linguistics\\
       School of Humanities, Languages and Social Sciences\\
       University of Wolverhampton\\
       Stafford St., Wolverhampton, WV1 1SB\\
       United Kingdom}

\title{NP Animacy Identification for Anaphora Resolution}

\maketitle

\begin{abstract}
In anaphora resolution for English, animacy identification can play
an integral role in the application of agreement restrictions
between pronouns and candidates, and as a result, can improve the
accuracy of anaphora resolution systems. In this paper, two methods
for animacy identification are proposed and evaluated using
intrinsic and extrinsic measures. The first method is a rule-based
one which uses information about the unique beginners in WordNet to
classify NPs on the basis of their animacy. The second method relies
on a machine learning algorithm which exploits a WordNet enriched
with animacy information for each sense. The effect of word sense
disambiguation on the two methods is also assessed. The intrinsic
evaluation reveals that the machine learning method reaches human
levels of performance. The extrinsic evaluation demonstrates that
animacy identification can be beneficial in anaphora resolution,
especially in the cases where animate entities are identified with
high precision.
\end{abstract}

%Keywords: \emph{animacy recognition, anaphora resolution, machine
%learning, evaluation}

%\bibliographystyle{fullname}

%\doublespace

\section{Introduction}

Anaphora resolution is the process which attempts to determine the
meaning of expressions such as pronouns or definite descriptions
whose interpretation depends on previously mentioned entities or
discourse segments. Anaphora resolution is very important in many
fields of computational linguistics such as machine translation,
natural language understanding, information extraction and text
generation \cite{mitkov-02}.

Previous work in anaphora resolution (AR) has shown that its levels
of performance are related to both the type of text being processed
and to the average number of noun phrases (NPs) under consideration
as a pronoun's antecedent \cite{evans-DAARC-00}. Acknowledging this,
researchers have proposed and incorporated various methods intended
to reduce the number of candidate NPs considered by their anaphora
resolution systems. Most approaches to pronominal anaphora
resolution rely on compatibility of the agreement features between
pronouns and antecedents, as a means of minimising the number of NP
candidates. Although, as noted by Barlow \citeyear{barlow-DAARC-98}
and Barbu, Evans, and Mitkov \citeyear{barbu-LREC-02}, this
assumption does not always hold, it is reliable in enough cases to
be of great practical value in anaphora and coreference resolution
systems. Such systems rely on knowledge about the number and gender
of NP candidates in order to check the compatibility between
pronouns and candidates
\cite{hobbs-76,lappin-94,kennedy-COLING-96,mitkov-98b,cardie-99,ng-02}.
In addition to number and gender compatibility, researchers reduced
the number of competing candidates considered by their systems by
means of syntactic filters \cite{hobbs-76,lappin-94}, semantic
filters \cite{hobbs-78} or discourse structure
\cite{brennan-ACL-87,cristea-COLING-00}.

In English, the automatic identification of the specific gender of
NPs is a difficult task of arguably limited utility. Despite this,
numerous researchers \cite{hale-98,denber-98,cardie-99} have
proposed automatic methods for identifying the potential gender of
NPs' referents. In this paper, the problem of \emph{animacy
identification} is tackled. The concern with animacy as opposed to
gender arises from the observation that animacy serves as a more
reliable basis for agreement between pronouns and candidates (see
examples in Section \ref{sec:what}). Animacy identification can be
very useful in tasks like anaphora resolution and coreference
resolution where the level of ambiguity can be reduced by filtering
out candidates which do not have the same value for animacy as the
anaphor, as well as in question answering, where it can be used to
improve system responses to ``who'' questions by allowing them to
ensure that the generated answers consist of animate references.

In this research, a NP is considered to be animate if its referent
can also be referred to using one of the pronouns \emph{he},
\emph{she}, \emph{him}, \emph{her}, \emph{his}, \emph{hers},
\emph{himself}, \emph{herself}, or a combination of such pronouns
(e.g. \emph{his/her}). Section \ref{sec:what} provides more clarity
on this definition, considering a range of exceptions and
problematic cases, as well as examining some consequences of this
treatment of animacy. The corpus used in this research is described
in Section \ref{sec:corpus}. In this paper several methods for
animacy identification are proposed and evaluated. First, a simple
statistical method based on WordNet \cite{fellbaum-98} is described
in Section \ref{sec:simple}. Following from the description of the
simple statistical method, a machine learning method that overcomes
some of the problems of the simple method, offering improved
performance, is described in Section \ref{sec:ML}. In the latest
stages of development, word sense disambiguation (WSD) is added to
further improve the accuracy of the classification. This is
presented in Section \ref{sec:WSD}. In Section \ref{sec:evaluation},
the systems are evaluated using both intrinsic and extrinsic
evaluation methods, and it is noted that the machine learning
methods reach human performance levels. Finally, Section
\ref{sec:related} is dedicated to related work and is followed by
conclusions.

\section{What Constitutes an Animate Noun Phrase?}\label{sec:what}

It has been argued that ``in English nouns are not classified
grammatically, but semantically according to their coreferential
relations with personal, reflexive and \emph{wh-}pronouns'' \cite[p.
314]{quirk-85}. According to their classification, animate noun
phrases contain both personal (e.g. male, female, dual, common and
collective nouns) and non-personal noun phrases (e.g. common,
collective and animal nouns). In this paper, our goal is to design a
method which improves the performance of anaphora resolution methods
by filtering out candidates which do not agree in terms of animacy
with a given referential pronoun. For this reason, the more specific
definition of animacy given in the introduction is used. This means
that in this paper, those noun phrases which can normally be
referred to by the pronouns \emph{he} and \emph{she} and their
possessive and reflexive forms, are considered animate, but no
distinction is made between those pronouns to determine their
gender. This view is adopted because, in the linguistic processing
of English documents, it is vital to distinguish between neuter and
animate references but problematic, and often of limited utility, to
distinguish between masculine and feminine ones.

To illustrate, in the sentence \emph{The primary user of the machine
should select his or her own settings}, considering the noun phrase
\emph{the primary user of the machine} to be either masculine or
feminine, and then applying strict agreement constraints between
this reference and the subsequent pronominal ones, will adversely
affect the performance of reference resolution systems because such
constraints will eliminate the antecedent from the list of
candidates of one of the pronouns depending on the gender attached
to the NP. Ideally, the reference should be considered animate -
compatible, in terms of agreement, with subsequent animate pronouns,
and incompatible with neuter pronouns.

\begin{figure*}[t]
\begin{center}
\includegraphics[width=12cm]{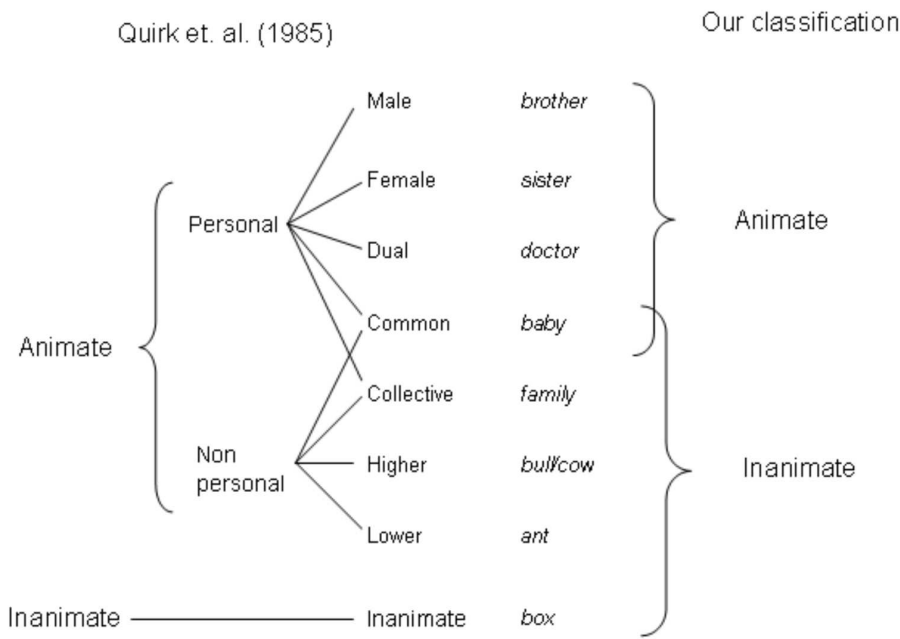}
\end{center}
\caption{Quirk el. al. (1985) vs our classification of animacy.
(Adapted from Quirk el. al. (1985, p. 314, Fig. 5.104))}
\label{fig:classification}
\end{figure*}

Figure \ref{fig:classification} presents the differences between
Quirk et. al.'s (1985) classification of animacy and the one used in
this paper. As can be seen in the figure, their
definition of animate nouns is much wider than that used in this
paper. We consider of the classes presented by Quirk et al.'s
(1985), the only animate entities to be \emph{male}, \emph{female},
\emph{dual}\footnote{Dual nouns are nouns that refer to people but
whose gender is underspecified such as \emph{artist, cook}, etc.}
and some \emph{common gender}\footnote{In Figure
\ref{fig:classification} these nouns are labeled as \emph{common}.}
nouns.

\emph{Common gender nouns} are defined by Quirk et al.
\citeyear{quirk-85} as intermediate between personal and
non-personal nouns. For us, their animacy can be either animate or
inanimate, depending on which pronoun is used to refer to them.
However, the animacy of some of the common nouns such as \emph{cat}
depends upon the perception of that entity by the speaker/writer. If
the noun is being used to refer to a pet, the speaker/writer is also
likely to use animate pronouns rather than inanimate ones to refer
to it. Such circumstances are not detected by our method: they may
be the focus of methods which try to identify the sentiments of
speakers/writers towards entities.

\emph{Collective nouns} such as \emph{team}, that refer to sets of
animate entities, may intuitively be considered animate. However,
the only suitable pronominal references for the denotation of such
phrases are singular neuter pronouns or plural pronouns of
unspecified gender. These referents are never referred to using
animate pronouns. Given that the \emph{raison d'etre} for our
research into animacy identification is the facilitation of
real-world anaphora resolution, such NPs are considered inanimate in
the current work, and not animate as they are by Quirk et al.
\citeyear{quirk-85}.

Collective nouns such as \emph{people} pose further problems to
annotation and processing. In some contexts the word \emph{people}
can be used as the plural form of \emph{person}, in which case it
should be considered animate by the definition presented earlier.
However in some cases it is used more generically to refer to
national populations (e.g. \emph{the peoples of Asia}) in which case
it should be considered inanimate. In light of this, it seems that
the class of this word depends on its context. However, in practical
terms, the morpho-syntactic parsing software that we use
\cite{tapanainen-97} returns \emph{people} and not \emph{person} as
the root the noun, so for this reason, the noun \emph{people} is
considered inanimate for our purposes. The same reasoning was
applied to other similar nouns. The drawback of this approach is
that annotators did not find this very intuitive and as a result
errors were introduced in the annotation (as discussed in the next
section).

The rest of the categories introduced by Quirk et al.
\citeyear{quirk-85}: \emph{non-personal higher}, \emph{non-personal
lower} and \emph{inanimate} correspond in our definition to
inanimate nouns. As with common gender nouns, it is possible to have
non-personal higher and lower nouns such as \emph{horse} and
\emph{rabbit} which can be pets and therefore are referred to by
speakers using \emph{he} or \emph{she}. As we cannot detect such
usages, they are all considered inanimate.\footnote{Actually on the
basis of the explanation provided by Quirk et al.
\citeyear{quirk-85} the distinction between common nouns and higher
and lower non-personal nouns when the latter are `personified' seems
very fuzzy.}

In the present work, the animacy of a noun phrase (NP) is considered
to derive from the animacy of its head. To illustrate, both
\emph{the man} and \emph{the dead man} can be referred to using the
same animate pronoun. Moreover, when considering the animacy of
plural NPs such as \emph{mileage claimants}, the singular form
\emph{mileage claimant} is derived and used as the basis of
classification because the plural form shares the animacy of its
singular form. In this way, our treatment of NP animacy mirrors the
treatment of grammatical number under the Government and Binding
Theory \cite{chomsky-81}. Under this approach, the projection
principle implies that agreement information for a NP is derived
from that of its head.

In this paper, the animacy of only common nouns is determined and
not of proper nouns such as named entities (NE). The reason for this
is that the separate task of named entity recognition is normally
used to classify NEs into different categories such as
\textsc{person}, \textsc{organization}, and \textsc{location}. Given
that they label entities of similar semantic types, these categories
can then be used to determine the animacy of all the entities that
belong to them. It is acknowledged that named entity recognition is
an important component in the identification of animate references,
but one which lies beyond the scope of the present work. Methods
based on semantics, such as the ones described in Section
\ref{sec:methods} are especially vulnerable to errors caused by a
failure to recognise the difference between words such as \emph{Cat}
or \emph{Bob} when used as common nouns which are inanimate
references or as proper nouns which are animate references.

\section{Corpus-Based Investigation}\label{sec:corpus}

The identification of NP animacy, as described in the previous
section, was amenable to a corpus-based solution. In this research
two corpora are being used: The first is a collection of texts from
Amnesty International (AI) which were selected because they contain
a relatively large proportion of references to animate entities. The
second is a selection of texts from the SEMCOR corpus
\cite{landes-98}, chosen because their nouns were annotated with
senses from WordNet. This annotation made them suitable for
exploitation in the development of the automatic method for animacy
identification described in Section \ref{sec:ML}. The SEMCOR corpus
was built on the basis of Brown Corpus \cite{frances-82} and for our
experiments we use texts from newswire, science, fiction and humor.

In order to make the data suitable for evaluation purposes, NPs from
the two corpora have been manually annotated with information about
their animacy. The characteristics of these corpora are summarized
in Table \ref{tab:char_corpora}. As can be seen in the table, even
though texts which contain many references to animate entities were
selected, the number of inanimate entities is still much larger than
the number of animate ones.

\begin{table}
\centering
\begin{footnotesize}
\begin{tabular}{|l|c|c|}
\hline & SEMCOR & AI \\\hline No of words & 104,612 & 15,767\\\hline
No of animate entities & 2,321 & 538\\\hline No of inanimate
entities & 17,380 & 2,586\\\hline Percentage of animate entities &
12\% & 21\% \\\hline Total entities & 19,701 & 3,124\\\hline
\end{tabular}
\end{footnotesize}
\caption{The characteristics of the two corpora
used}\label{tab:char_corpora}
\end{table}

To assess the difficulty of the annotation task, and implicitly, to
estimate the upper performance limit of automatic methods, a second
annotator was asked to annotate a part of the corpus and
inter-annotator agreement was calculated. To this end, the whole AI
corpus, and nine texts with over 3,500 references from the SEMCOR
corpus have been randomly selected and annotated. Comparison between
the two annotations revealed a level of agreement of 97.5\% between
the two annotators and a value of 0.91 for the kappa statistic which
indicates very high agreement between annotators. The agreement on
the SEMCOR data was slightly higher than that for the AI corpus, but
the difference was not statistically significant.

Investigation of the annotation performed by the two annotators and
discussion with them revealed that the main source of disagreement
was the monotony of the task. The two annotators had to use a simple
interface which displayed for each sentence one NP at a time, and
were required to indicate whether the NP was animate or inanimate by
choosing one of two key strokes. Due to the large number of
inanimate entities in the corpus, the annotators often marked
animate entities as inanimate accidentally. In some cases they
noticed their mistake and corrected it, but it is very likely that
many such mistakes went unobserved.

Another source of disagreement were collective nouns such as
\emph{people}, \emph{government}, \emph{jury} or \emph{folk} which
according to the discussion in Section \ref{sec:what} should
normally be marked as inanimate. In some cases, the context of the
NP or tiredness on the part of the annotator led to them being
erroneously marked as animate. Similarly, it was noticed that the
annotators wrongly considered some plural noun phrases such as
\emph{observers, delegates, communists,} and \emph{assistants} to be
collective ones and marked them as inanimate. However, it is likely
that some of these errors were introduced due to the monotony of the
task. Unfamiliar nouns such as \emph{thuggee}, and words used in
some specialized domains such as baseball also caused difficulties.
Finally, another source of error arose from the use of Connexor's
FDG Parser \cite{tapanainen-97} to identify the noun phrases for
annotation. As a result, some of the noun phrases recognized by the
system were ambiguous (e.g. \emph{specialists and busy people} was
presented as one NP and according to the definition of animacy
adopted in the present work, \emph{specialists} is animate, whereas
\emph{busy people} is inanimate\footnote{It can be argued that the
singular form of \emph{people} is \emph{person}, and that it should
therefore be marked as animate. However, as discussed in Section
\ref{sec:what} due to the way it is processed by the preprocessing
tools employed here, annotators were asked to consider it
inanimate}.).

\section{Methods for Animacy Identification}\label{sec:methods}

By contrast to the situation with proper name recognition and
classification, which can exploit surface textual clues such as
capitalization and the explicit occurrence of words in a small
gazetteer of titles, knowledge as to the animacy of common NPs
appears to be purely implicit. Recognition of references to animate
entities must, at some point, be grounded in world-knowledge and
computed from explicit features of the text. This section presents
two methods developed for animacy identification which rely on
information extracted from WordNet, an electronic lexical resource
organized hierarchically by relations between sets of synonyms or
near-synonyms called synsets \cite{fellbaum-98}. The first method is
a rule-based one which employs a limited number of resources and is
presented in Section \ref{sec:simple}. Its shortcomings are
addressed by the machine learning method presented in Section
\ref{sec:ML}. Both methods consider all the senses of a word before
taking a decision about its animacy. For this reason, the word sense
disambiguation (WSD) module briefly discussed in Section
\ref{sec:WSD} was integrated into them.

\subsection{Rule-Based Method}\label{sec:simple}

In WordNet, each of the four primary classes of content-words
(nouns, verbs, adjectives and adverbs) are arranged under a small
set of top-level hypernyms called unique beginners
\cite{fellbaum-98}. Investigation of these unique beginners revealed
that several of them were of interest with respect to the aim of
identifying the animate entities in a text. In the case of nouns
there are 25 unique beginners, three of which are expected to be
hypernyms of senses of nouns that usually refer to animate entities.
These are \emph{animal}, reference number (05), \emph{person} (18),
and \emph{relation} (24).\footnote{The unique beginner \emph{animal}
corresponds to both animate and inanimate entities while
\emph{relation} subsumes mainly human relationships such as
\emph{brother}, \emph{sister}, \emph{parent}, etc.} There are also
four verb sense hierarchies out of fourteen, that allow the
inference to be made that their subject NPs should be animate. The
unique beginners in these cases are \emph{cognition} (31),
\emph{communication} (32), \emph{emotion} (37) and \emph{social}
(41).\footnote{The \emph{social} unique beginner subsumes relations
such as \emph{abdicate}, \emph{educate} and \emph{socialize}.} It
has been noted that inanimate entities such as organizations and
animals can also be agents of these types of verb, but it is
expected in the general case that these instances will be rare
enough to ignore. In light of the way in which WordNet is organized,
it was clear that it could be exploited in order to associate the
heads of noun phrases with a measure of confidence that the
associated NP has either an animate or inanimate referent.

It is very common for a noun to have more than one meaning, in many
cases corresponding to sense hierarchies which start from different
unique beginners. For this reason, the decision about whether a noun
phrase is animate or inanimate should be taken only after all the
possible senses of the head noun have been consulted. Given that
some of these senses are animate whilst others are inanimate, an
algorithm which counts the number of animate senses that are listed
for a noun (hyponyms of unique beginners 05, 18, or 24) and the
number of inanimate senses (hyponyms of the remaining unique
beginners) was proposed. Two ratios are then computed for each noun:
$$Noun\;animacy\;(NA)=\frac{Number\;of\;animate\;senses}{Total\;number\;of\;senses}$$
$$Noun\;inanimacy\;(NI)=\frac{Number\;of\;inanimate\;senses}{Total\;number\;of\;senses}$$
and compared to pre-defined thresholds in order to classify them as
animate or inanimate. Similarly, in the case of nouns that are the
heads of subject NPs, counts are made of the animate and inanimate
senses of the verbs they are subjects of and used to calculate
\emph{Verb animacy (VA)} and \emph{Verb inanimacy (VI)} in the same
way as \emph{NA} and \emph{NI}. These ratios are also used to
determine the animacy of the subject NP. Finally, contextual rules
(e.g. the presence of NP-internal complementizers and reflexives
such as \emph{who} or \emph{herself}) are applied in order to
improve the classification. The algorithm is presented in Algorithm
\ref{alg:simple-method} and evaluated in Section
\ref{sec:evaluation}. The three thresholds used in the algorithm
were determined through experimentation and the best values were
found to be $t_1=0.71$, $t_2=0.92$ and $t_3=0.90$.

\begin{algorithm}[t]
  \SetLine
  \KwData{NP is the noun phrase for which animacy has to be determined, $t_1$, $t_2$, $t_3$}
  \KwResult{The animacy of the NP}
  Compute NA, NI, VA, VI for NP\;
  \If{$NA > t_1$} {
    NP if animate\;
    Stop\;
  }
  \If{$NI > t_2$} {
    NP is inanimate\;
    Stop;
  }
  \If{($NA > NI$) and ($VA > VI$)} {
    NP is animate\;
    Stop\;
  }
  \If{(NP contains the complementizer \emph{who}) or ($VA > t_3$)} {
    NP is animate\;
    Stop\;
  }
  NP is inanimate\;
  \caption{The rule-based algorithm used to determine the animacy of a
noun phrase} \label{alg:simple-method}
\end{algorithm}

\subsection{Machine Learning for Animacy Identification}\label{sec:ML}

The method presented in the previous section has two main
weaknesses. The first one is that the unique beginners used to
determine the number of animate/inanimate senses are too general,
and in most cases they do not reliably indicate the animacy of each
sense in the class. The second weakness is due to the na\"{i}ve
nature of the rules that decide whether a NP is animate or not.
Their application is simple and involves a comparison of values
obtained for a NP with threshold values that were determined on the
basis of a relatively small number of experiments. In light of these
problems, a two step approach, each addressing one of the
aforementioned weaknesses, was proposed. In the first step, an
annotated corpus is used to determine the animacy of WordNet
synsets. This process is presented in Section \ref{sec:ML-step1}.
Once this information is propagated through the whole of WordNet, it
is used by a machine learning algorithm to determine the animacy of
NPs. This method is presented in Section \ref{sec:ML-step2}.

\subsubsection{The Classification of the Senses}\label{sec:ML-step1}

As previously mentioned, the unique beginners are too general to be
satisfactorily classified as wholly animate or inanimate. However,
this does not mean that it is not possible to uniquely classify more
specific senses as animate or inanimate. In this section, a
corpus-based method which classifies synsets from WordNet according
to their animacy is presented.

The starting point for classifying the synsets was the information
present in our annotated version of the SEMCOR corpus. The reason
for this is that by adding our animacy annotation to nouns which
were annotated with their corresponding sense from WordNet, this
information could be used to determine the animacy of the synset.
However, due to linguistic ambiguities and tagging errors not all
the senses can be classified adequately in this way. Moreover, many
senses from WordNet do not appear in SEMCOR, which means that no
direct animacy information can be determined for them. In order to
address this problem, the decision was made to use a bottom up
procedure which begins by classifying unambiguous terminal nodes and
then propagates this information to more general nodes. A terminal
node is unambiguously classified using the information from the
annotated files if all its occurrences in the corpus are annotated
with the same class. In the same way, a more general node can be
unambiguously classified if all of its hyponyms have been assigned
to the same class.

Due to annotation errors or rare uses of a sense, this condition is
rarely met and a statistical measure must be employed in order to
test the animacy of a more general node. A simple approach which
classifies a synset using a simple voting procedure on behalf of its
hyponyms will be unsatisfactory because it is necessary to know when
a node is too general to be able to assign it to one of the classes.
For this reason a statistical measure was used to determine the
animacy of a node in ambiguous cases.

The statistical measure used in this process is chi-squared, a
non-parametric test which can be used to estimate whether or not
there is any significant difference between two different
populations. In order to test whether or not a node is animate, the
two populations to be compared are:

\begin{enumerate}
\item  an observed population which
consists of the senses of the node's hyponyms which were annotated
as animate, and
\item  a hypothetical population in which all of the node's
hyponyms are animate.
\end{enumerate}

\begin{figure*}[t]
\begin{center}
\includegraphics[width=12cm]{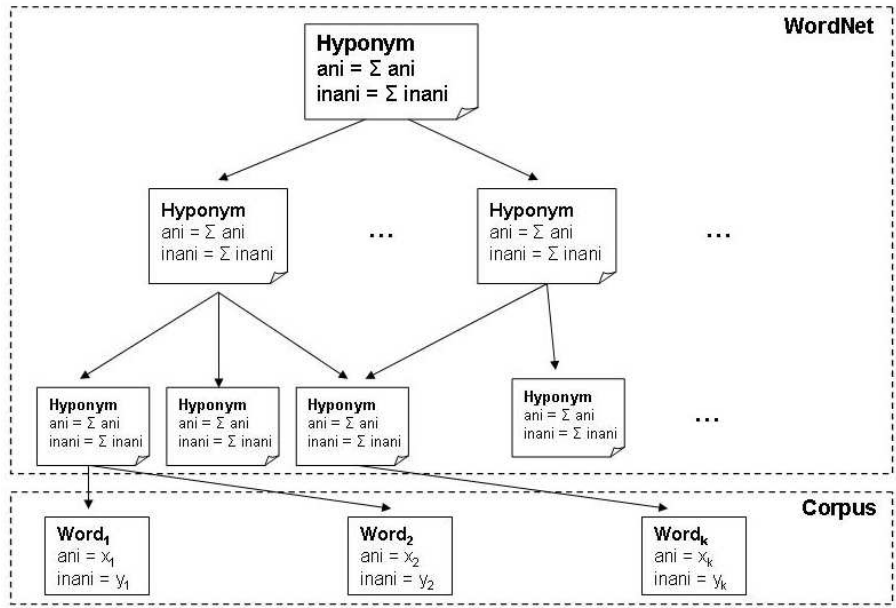}
\end{center}
\caption{Example showing the propagation of animacy from the corpus
to more general senses} \label{fig:tree}
\end{figure*}

If chi-square indicates that there is no difference between the two
populations then the node is classified as animate. The same process
is repeated in order to classify an inanimate node. If neither test
is passed, it means that the node is too general, and it and all of
its hypernyms can equally refer to both animate and inanimate
entities. In unambiguous cases (i.e. when all the hyponyms observed
in the corpus\footnote{Either directly or indirectly via hyponymy
relations.} are annotated as either animate or inanimate, but not
both), the more general node is classified as its hyponyms are. The
way in which information is propagated from the corpus into WordNet
is presented in Figure \ref{fig:tree}.

\begin{table*}[t]
\begin{center}
\begin{footnotesize}
\begin{tabular}{|c|c|c|c|c|c|}
\hline & $Sense_1$ & $Sense_2$ & $Sense_3$ & ... & $Sense_n$\\
\hline Observed & $ani_1$ & $ani_2$ & $ani_3$ & ... & $ani_n$\\
\hline Expected & $ani_1+inani_1$ & $ani_2+inani_2$ &
$ani_3+inani_3$ & ... & $ani_n+inani_n$ \\\hline \end{tabular}
\caption{Contingency table for testing the animacy of a
hypernym}\label{tab:conti}
\end{footnotesize} \end{center}
\end{table*}

To illustrate, for a more general node which has $n$ hyponyms the
contingency table (Table \ref{tab:conti}) can be built and used to
determine its animacy. Each hyponym is considered to have two
attributes: the number of times it has been annotated as animate
($ani_i$) and the number of times it has been annotated as inanimate
($inani_i$). The figures for $ani_i$ and $inani_i$ include both the
number of times that the sense directly appears in the corpus and
the number of times it appears indirectly via its hyponyms. Given
that the system is testing to see whether the more general node is
animate or not, for each of its hyponyms, the total number of
occurrences of a sense in the annotated corpus is the \emph{expected
value} (meaning that all the instances should be animate and those
which are not marked as animate are marked that way because of
annotation error or rare usage of the sense) and the number of times
the hyponym is annotated as referring to an animate entity is the
\emph{observed value}. Chi-square is calculated, and the result is
compared with the critical level obtained for $n-1$ degrees of
freedom and a significance level of .05. If the test is passed, the
more general node is classified as animate.

In order to be a valid test of significance, chi-square usually
requires expected frequencies to be 5 or more. If the contingency
table is larger than two-by-two, some few exceptions are allowed as
long as no expected frequency is less than one and no more than 20\%
of the expected frequencies are less than 5 \cite{sirkin-95}. In the
present case it is not possible for expected frequencies to be less
than one because this would entail no presence in the corpus. If,
when the test is applied, more than 20\% of the senses have an
expected frequency less than 5, the two similar senses with the
lowest frequency are merged and the test is repeated.\footnote{In
this context, two senses are considered similar if they both have
the same attribute (i.e. animacy or inanimacy) equal to zero.} If no
senses can be merged and still more than 20\% of the expected
frequencies are less than 5, the test is rejected.

This approach is used to classify all the nodes from WordNet as
animate, inanimate or undecided. The same approach is also employed
to classify the animacy of verbs on the basis of the animacy of
their subjects. An assessment of the coverage provided by the method
revealed that almost 94\% of the nodes from WordNet can be
classified as animate or inanimate. This is mainly due to the fact
that some very general nodes such as \emph{person}, \emph{plant} or
\emph{abstraction} can be classified without ambiguity and as a
result all their hyponyms can be classified in the same way. This
enriched version of WordNet is then used to classify nouns as
described in the next section.

\subsubsection{The Classification of a Noun}\label{sec:ML-step2}

The classification described in the previous section is useful for
determining the animacy of a sense, even for those which were not
previously found in the annotated corpus, but which are hyponyms of
a node that has been classified. However, nouns whose sense is
unknown cannot be classified directly and therefore an additional
level of processing is necessary. In this section, the use of
{\sc t}i{\sc mbl} \cite{TiMBL-00} to determine the animacy of nouns is
described.

{\sc T}i{\sc mbl} is a program which implements several machine
learning techniques. Experimenting with the algorithms available in
{\sc t}i{\sc mbl} with different configurations, the best results were
obtained using instance-based learning with gain ratio as the
weighting measure \cite{quinlan-93,mitchell-97}. In this type of
learning, all the instances are stored without trying to infer
anything from them. At the classification stage, the algorithm
compares a previously unseen instance with all the data stored at
the training stage. The most frequent class in the k nearest
neighbors is assigned as the class to which that instance belongs.
After experimentation, it was noticed that the best results were
obtained when the three nearest neighbors were used (k=3), the
distance between two instances is calculated using overlap metric
and the importance of each feature is weighted using gain ratio
\cite{TiMBL-00}.

In the present case, the instances used in training and
classification consist of the following information:
\begin{enumerate}
\item The lemma of the noun which is to be classified.
\item The number of animate and inanimate senses of the word.
As mentioned before, in the cases where the animacy of a sense is
not known, it is inferred from its hypernyms. If this information
cannot be found for any of a word's hypernyms, information on the
unique beginners for the word's sense is used, in a manner similar
to that used by the rule-based system described in Section
\ref{sec:simple}.
\item For the heads of subject NPs, the number of
animate/inanimate senses of its verb. For those senses for which the
classification is not known, an algorithm similar to the one
described for nouns is employed. These values are 0 for heads of
non-subjects.
\item The ratio of the number of animate singular pronouns
(e.g \emph{he} or \emph{she}) to inanimate singular pronouns (e.g.
\emph{it}) in the whole text. The justification for this feature is
that a text containing a large number of gender marked pronouns will
be more likely to mention many animate entities
\end{enumerate}

These features were encoded as vectors to be classified by
{\sc t}i{\sc mbl} using the algorithm and settings described earlier.
The algorithm described in this section is evaluated in Section
\ref{sec:evaluation}.

\subsection{Word Sense Disambiguation}\label{sec:WSD}

It is difficult to disambiguate the possible senses of words in
unrestricted texts, but it is not so difficult to identify those
senses which are more likely to be used in a text than others. Such
information was not considered in the methods presented in Sections 
\ref{sec:simple} and \ref{sec:ML}. Instead, in those methods, all
the senses were considered to have an equal weight. In order to
address this shortcoming, the word sense disambiguation (WSD) method
described by Resnik \citeyear{resnik-95} was implemented and used in
the classification algorithm. The WSD method computes the weight of
each possible sense of each noun by considering the other nouns in a
text. These weights were used to compute the number of
animate/inanimate senses. The underlying hypothesis is that the
animacy/inanimacy of senses which are more likely to be used in a
particular text should count more than that of improbable senses.
The impact of this approach on the animacy identifiers presented in
the previous section is also evaluated.

\section{Evaluation of the Systems}\label{sec:evaluation}

In this section, the systems presented in Section \ref{sec:methods}
are evaluated using \emph{intrinsic} and \emph{extrinsic} evaluation
methods \cite{sparck-96}. Both evaluation methods are necessary
because the aim is not only to find out which of the methods can
classify references to animate entities most accurately, but also to
assess how appropriate they are for inclusion into an anaphora
resolution method. In addition, the complexity of the systems is
considered.

In order to increase the reliability of the evaluation, the systems
are assessed on both corpora described in Section \ref{sec:corpus}.
The thresholds used in the simple method presented in Section
\ref{sec:simple} were determined through direct observation of the
performance results when the system was applied to the AI corpus.
Evaluating the method on the SEMCOR corpus allows its performance to
be measured on completely unseen data. In addition, the texts from
SEMCOR are in a completely different genre from AI, allowing an
assessment to be made of the degree to which the system described in
Section \ref{sec:simple} is genre independent.

Evaluation raises more serious problems when the machine learning
method is considered. As is well known, whenever a machine learning
method is evaluated, a clear distinction has to be made between
training data and testing data. In the case of the system described
in Section \ref{sec:ML}, the approach was evaluated using 10-fold
cross-validation over the SEMCOR corpus. Given that the AI corpus is
available, the systems can also be evaluated on data from a domain
which was not used in setting the parameters of the machine learning
method. In addition, the evaluation of the machine learning methods
on the AI corpus is useful in proving that the classification of the
synsets from WordNet on the basis of the animacy annotation added to
SEMCOR can be used to develop a system whose performance is not
text-dependent.

\subsection{Intrinsic Evaluation}\label{intrinsic}

Intrinsic evaluation methods measure the accuracy of a system in
performing the task which it was designed to carry out. In the
present case, it is the accuracy with which an entity can be
classified as animate or inanimate. In order to assess the
performance of the systems, four measures are considered:

\begin{equation}\label{eq:accuracy}
Accuracy=\frac{Correctly\;classified\;items}{Total\;number\;of\;items}
\end{equation}
\begin{equation}\label{eq:precision}
Precision=\frac{True\;positives}{True\;positives\;+\;False\;positives}
\end{equation}
\begin{equation}\label{eq:recall}
Recall=\frac{True\;positives}{True\;positives\;+\;False\;negatives}
\end{equation}
\begin{equation}\label{eq:fmeasure}
F{-}measure=\frac{2*Precision*Recall}{Precision+Recall}
\end{equation}

\begin{figure*}[t]
\begin{center}
\includegraphics[width=12cm]{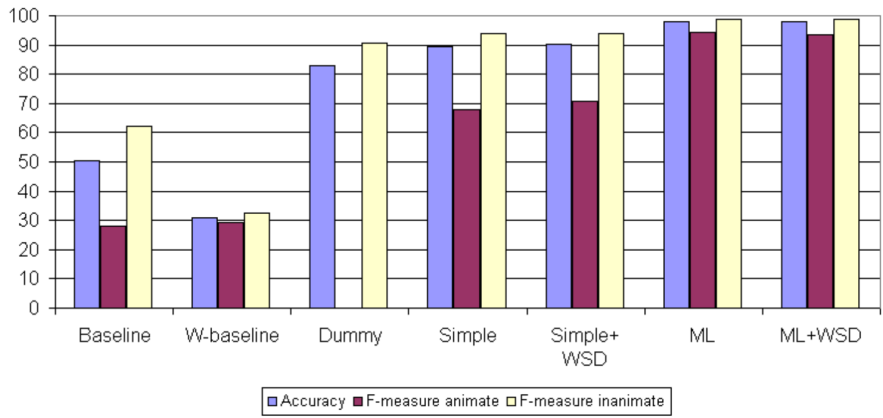}
\end{center}
\caption{Evaluation of methods on AI corpus} \label{fig:ie_AI}
\end{figure*}

\begin{figure*}[t]
\begin{center}
\includegraphics[width=12cm]{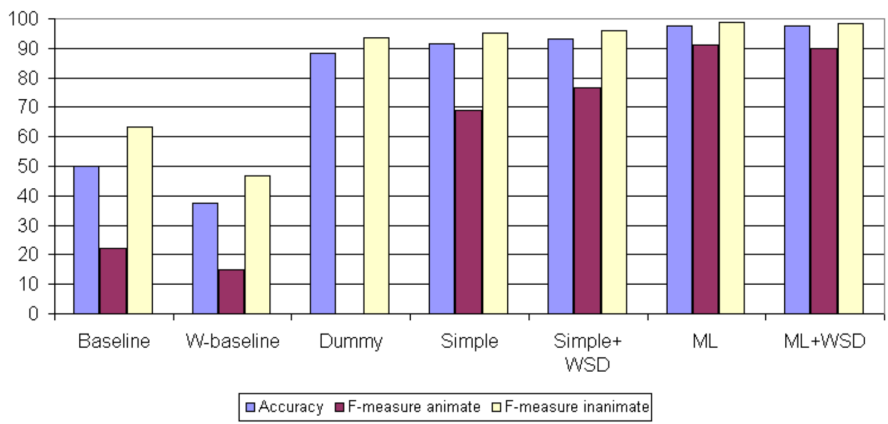}
\end{center}
\caption{Evaluation of methods on SEMCOR corpus}
\label{fig:ie_semcor}
\end{figure*}

The \emph{accuracy} (\ref{eq:accuracy}) measures how well a system
can correctly classify a reference to an entity as animate or
inanimate, but it can be misleading because of the large number of
inanimate entities mentioned in texts. As is clear from Table
\ref{tab:char_corpora}, even though the texts were chosen so as to
contain a large number of references to animate entities, the ratio
between the number of references to animate entities and inanimate
entities is approximatively 1 to 7.5 for SEMCOR, and 1 to 4.8 for
AI. This means that a method which classifies all references to
entities as inanimate would have an accuracy of 88.21\% on SEMCOR
and 82.77\% on AI. As can be seen in Figures \ref{fig:ie_AI} and
\ref{fig:ie_semcor}, as well as in Table \ref{tab:intrinsic-res},
these results are not very far from the accuracy obtained by the
system described in Section \ref{sec:simple}. However, as mentioned
earlier, the intention is to use the filtering of references to
animate entities for anaphora resolution and therefore, the use of a
filter which classifies all the references as inanimate would be
highly detrimental.

It is clearly important to know how well a system is able to
identify references to animate and inanimate entities. In order to
measure this, \emph{precision} (\ref{eq:precision}) and
\emph{recall} (\ref{eq:recall}) are used for each class. The
precision with which a system can identify animate references is
defined as the ratio between the number of references correctly
classified by the system as animate and the total number of
references it classifies as animate (including the wrongly
classified ones). A method's recall in classifying references to
animate entities is defined as the ratio between the number of
references correctly classified as animate and the total number of
animate references to be classified. The precision and recall of
inanimate classification is defined in a similar manner. The
\emph{f-measure} (\ref{eq:fmeasure}) combines precision and recall
into one value. Several formulae for f-measure were proposed, the
one used here gives equal importance to precision and recall.

Figures \ref{fig:ie_AI} and \ref{fig:ie_semcor}, as well as Table
\ref{tab:intrinsic-res} at the end of the paper, present the
accuracy of the classification, and f-measures for classifying the
animate and inanimate references. In addition to the methods
presented in Section \ref{sec:methods}, three baseline methods were
introduced. The first one classifies a reference to an entity as
animate or inanimate on a random basis and is referred to in the
figures as \emph{baseline}. A second random baseline was introduced
because it was assumed that the number of gender marked pronouns in
a text can indicate how likely it is that a particular noun
appearing in that text will be animate or inanimate. In this case,
the probability of a reference to be animate is proportional to the
number of animate pronouns in the text and the classification is
made on a weighted random basis. A similar rule applies for
inanimate references. This second baseline is referred to in the
figures as \emph{W-baseline}. For purposes of comparison, a method
was included which classifies all references as inanimate. This
method is referred to as the \emph{dummy method}.

Figures \ref{fig:ie_AI} and \ref{fig:ie_semcor} show that all the
other methods significantly outperform the baselines used. Close
investigation of the figures, as well as of Table
\ref{tab:intrinsic-res}, shows that, for both corpora, the best
method is the one which uses machine learning (presented in Section
\ref{sec:ML}). It obtains high accuracy when classifying references
to both animate and inanimate entities. In terms of accuracy, the
simple method performs unexpectedly well, but it fails to accurately
classify references to animate entities. Moreover, comparison with
the dummy method on both files shows that the results of the simple
method are not much better, which suggests that the simple method
has a bias towards recognition of references to inanimate entities.
The integration of word sense disambiguation yields mixed results:
it increases the accuracy of the simple method, but it slightly
decreases the performance of the machine learning method.

The relatively poor accuracy of the \emph{Simple system} was
expected and can be explained by the fact that the unique beginners,
which are used as the basis for classification in that method, cover
a range of senses which is too wide to be classified as belonging to
a single animate or inanimate class. They are too general to be used
as the basis for accurate classification. Additionally, the rules
used to assist classification only provide limited recall in
identifying animate references.

Comparison between the accuracy of the machine learning method and
the level of inter-annotator agreement reveals that the automatic
method agrees with the first annotator more than the second
annotator does. As a result of this, it can be concluded that the
accuracy of the automatic method matches human performance levels.

\subsection{Extrinsic evaluation}\label{sec:extrinsic}

In the previous section, the performance of the classification
methods was evaluated and it was demonstrated that even simple
methods can achieve high accuracy at the expense of low precision
and recall in the classification of references to animate entities.
In computational linguistics, the output of one method is often used
as the input for another one, and therefore it is important to know
how the results of the first method influence the results of the
second. This kind of evaluation is called \emph{extrinsic
evaluation}. Given that the identification of references to animate
entities is not very useful in its own right, but can be vital for
tasks like anaphora resolution, it is necessary to perform extrinsic
evaluation too. In the case of this evaluation, the assumption is
that the performance of anaphora resolution can be improved by
filtering out candidates which do not agree in animacy with each
referential pronoun.

\begin{figure*}[t]
\begin{center}
\includegraphics[width=10cm]{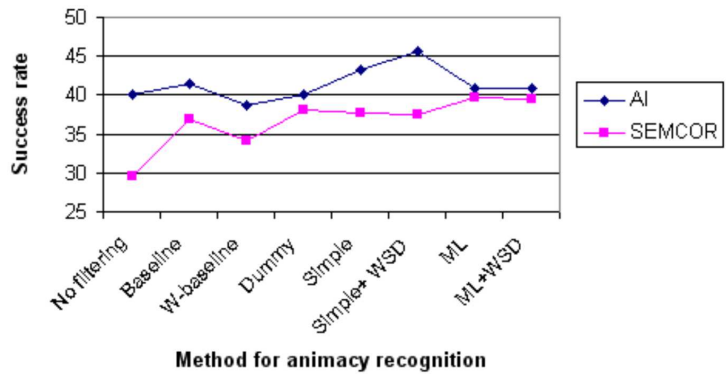}
\end{center}
\caption{The accuracy of {\sc mars} when the different animacy
filters are applied} \label{extrinsic-eval-MARS}
\end{figure*}

The influence of animacy identification on anaphora resolution is
thus evaluated using {\sc mars} \cite{mitkov-cicling-02}, a robust
anaphora resolver which relies on a set of boosting and impeding
indicators to select an antecedent from a set of competing
candidates. Due to the fact that the evaluation of {\sc mars}
requires the manual annotation of pronouns' antecedents, which is a
time consuming task, this evaluation was carried out only on a part
of the corpus. To this end, the entire Amnesty International corpus
as well as 22 files from the SEMCOR corpus have been used. Given
that the animacy identifier can only influence the accuracy of
anaphora resolvers with respect to third person singular pronouns,
the accuracy of the resolver is reported only for these pronouns.
Accuracy in anaphora resolution was calculated as the ratio between
the number of pronouns correctly resolved and the total number of
third person singular pronouns appearing in the test data. Figure
\ref{extrinsic-eval-MARS} and Table \ref{tab:extrinsic} display this
accuracy for alternate versions of {\sc mars} that exploit different
methods for animacy identification.

\begin{figure*}[t]
\begin{center}
\includegraphics[height=5cm]{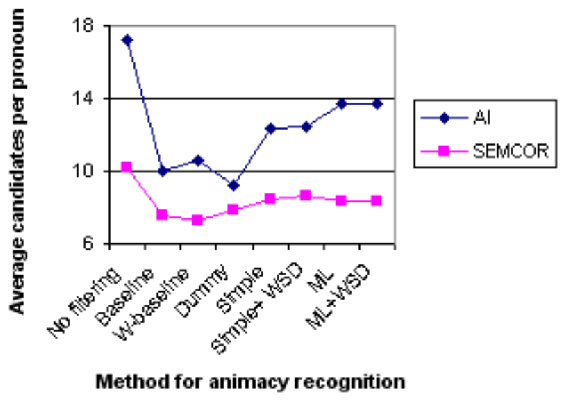}
\includegraphics[height=5cm]{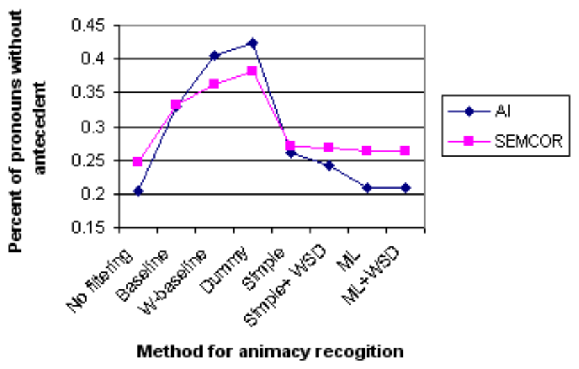}
\end{center}
\caption{The average number of candidates and the percentage of
pronouns without correct candidates when different animacy filters
are applied} \label{fig:extrinsic-eval-pc}
\end{figure*}

{\sc Mars} was designed to process texts from the technical domain,
and therefore its performance is rather poor on this test corpus.
Moreover, its performance can vary greatly from one domain to
another. In light of the fact that the results of a different
anaphora resolver may be very different on the same set of data, in
addition to the performance of {\sc mars} with respect to third
person singular pronouns, Figure \ref{fig:extrinsic-eval-pc} and
Table \ref{tab:extrinsic} also present the reduction in the number
of candidates that results from the animacy filtering, and the
increase in the number of pronouns whose sets of competing
candidates contain no valid antecedents as a result of this
filtering. The former number is presented as the average number of
candidates per pronoun, and the latter as the percentage of pronouns
without valid antecedents in the list of candidates. The
justification for reporting these two measures is that a good
animacy filter will eliminate as many candidates as possible, but
will not eliminate antecedents and leave pronouns without any
correct candidates to be resolved to.\footnote{It should be noted
that, even without filtering, there are pronouns which do not have
any candidates due to errors introduced by preprocessing tools such
as the NP extractor which fails to identify some of the NPs.}

As can be seen in Figure \ref{extrinsic-eval-MARS} and Table
\ref{tab:extrinsic}, regardless of which animacy identification
method is used, the accuracy of the anaphora resolver improves. The
degree of improvement varies from one corpus to another, but the
pattern regarding the reduction in the number of candidates and the
increase in the number of pronouns whose sets of competing
candidates contain no valid antecedent is the same across both
corpora. For the AI corpus, the best performance is obtained when
the simple method enhanced with word sense disambiguation is used,
followed by the simple method. Both improvements are statically
significant\footnote{In all the cases where we checked whether the
differences between two results are statistically significant we
used t-test with 0.05 confidence level.}, as well as the difference
between them. Both versions of the machine learning method improve
the success rate of {\sc mars} by a small margin which is not
statistically significant, but they increase the number of pronouns
with no valid antecedent to select by only one, an increase which is
not statistically significant. For the simple methods, the increase
in the number of this type of pronoun is much larger and is
statistically significant. Therefore in the case of the AI corpus,
it can be concluded that, for {\sc mars}, a more aggressive method
for filtering out candidates, such as the simple method with word
sense disambiguation, is more appropriate. However, it is possible
that for other anaphora resolution methods this result is not valid
because they may be more strongly influenced by the increase in the
number of pronouns with no valid antecedent to select.

Processing the SEMCOR corpus, the best results for {\sc mars} are
obtained by the machine learning method without the WSD module
followed by the one which performs WSD. In both cases the increase
over the performance of the unfiltered version is statistically
significant, but the differences between the two machine learning
methods are too small to be significant. In addition, these two
methods ensure a large reduction in the number of candidates with
the smallest increase in the number of pronouns whose sets of
competing candidates contain no valid antecedent, an increase which
is not significant.

As expected, the three baselines perform rather poorly. All three of
them reduce the number of candidates at the expense of a high
increase in the number of pronouns with no valid antecedent
available for selection. Both the reduction in the number of
candidates and the increase in the number of pronouns with no valid
antecedent are statistically significant when compared to the system
that does not use any filtering.

The results of {\sc mars}'s performance are rather mixed when these
baselines are used. For the AI corpus, the random baseline leads to
a better result for {\sc mars} than the machine learning methods, but
the differences are not statistically significant. However, this is
achieved with a large increase in the number of pronouns which
cannot be correctly resolved because all their valid antecedents
have also been filtered by the method. For the AI corpus,
application of the other two baselines led to results worse than or
equal to those of {\sc mars} when no filtering is applied as a result
of the large drop in the number of candidates.

For the SEMCOR corpus, all three baselines give rise to
statistically significant improvements in performance levels over
those obtained when no filtering is applied, but this is achieved by
dramatically reducing the number of candidates considered.
Integration of the dummy method into {\sc mars} leads to results
which are better than the simple methods but, as argued before, this
method is not appropriate for anaphora resolvers because it prevents
them from correctly resolving any animate pronoun.

Investigation of the results revealed that for about 31\% of the
candidates it was not possible to apply any animacy filtering. There
are three reasons for this. First, in the majority of cases,
candidates are named entities which, as mentioned in Section
\ref{sec:what}, are not tackled by our method, though they
constitute a relatively high proportion of the noun phrases
occurring in the chosen texts. A second reason for these cases is
the fact that some of the words are not present in WordNet and as a
result, they are ignored by our method. Finally, in a limited number
of cases the noun phrases identified by {\sc mars} did not match
those identified by our animacy identifiers and for this reason it
was not possible to classify them.\footnote{The animacy identifiers
proposed in this paper use both the NP and its context (i.e. the
verb on which it depends and the number of pronouns in the text) and
therefore they have to be run independently from any other module
which uses their results.}

\subsection{Extrinsic Evaluation on Simulated Data}

\begin{figure*}[t]
\begin{center}
\includegraphics[height=4cm]{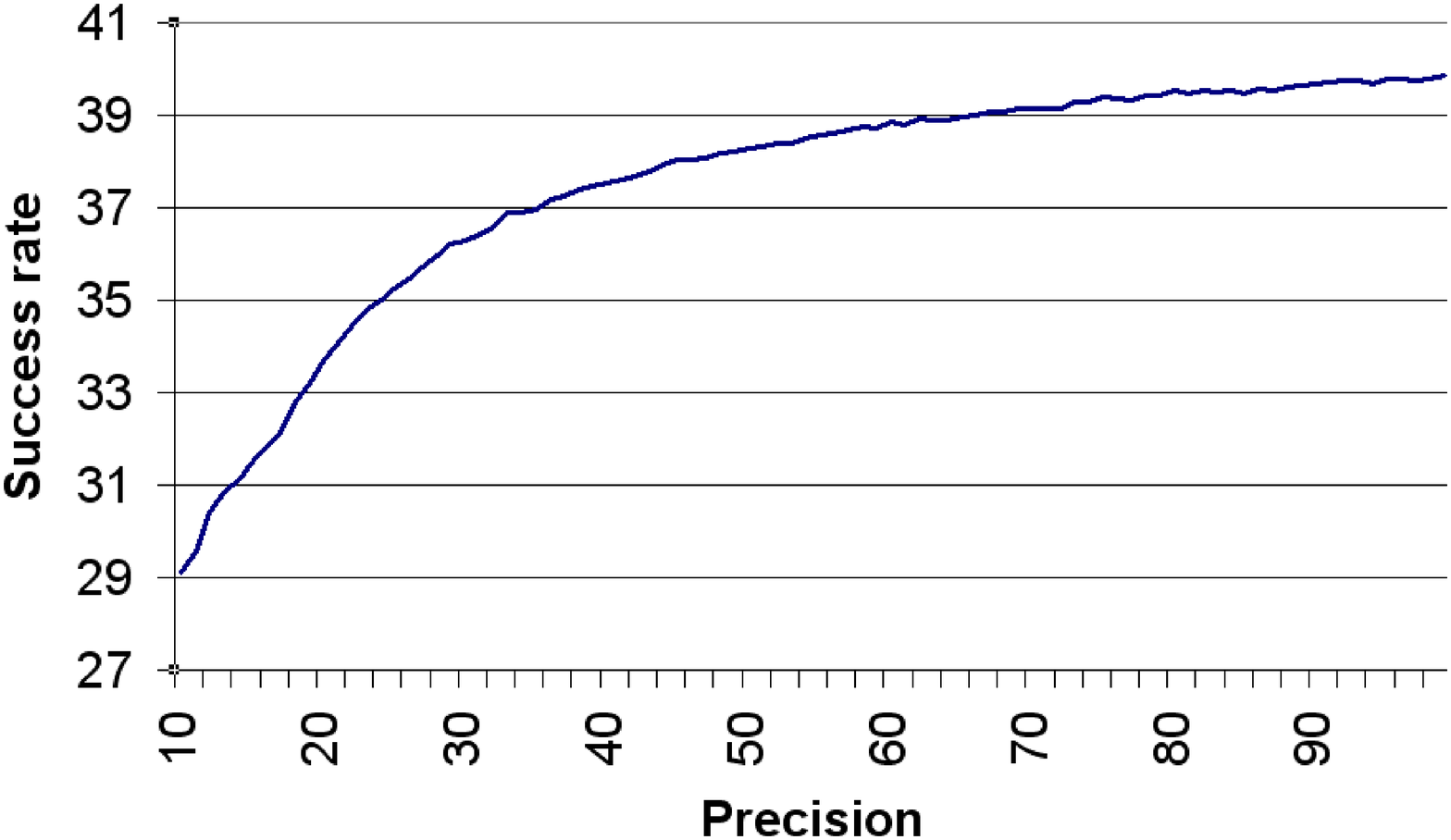}
\includegraphics[height=4cm]{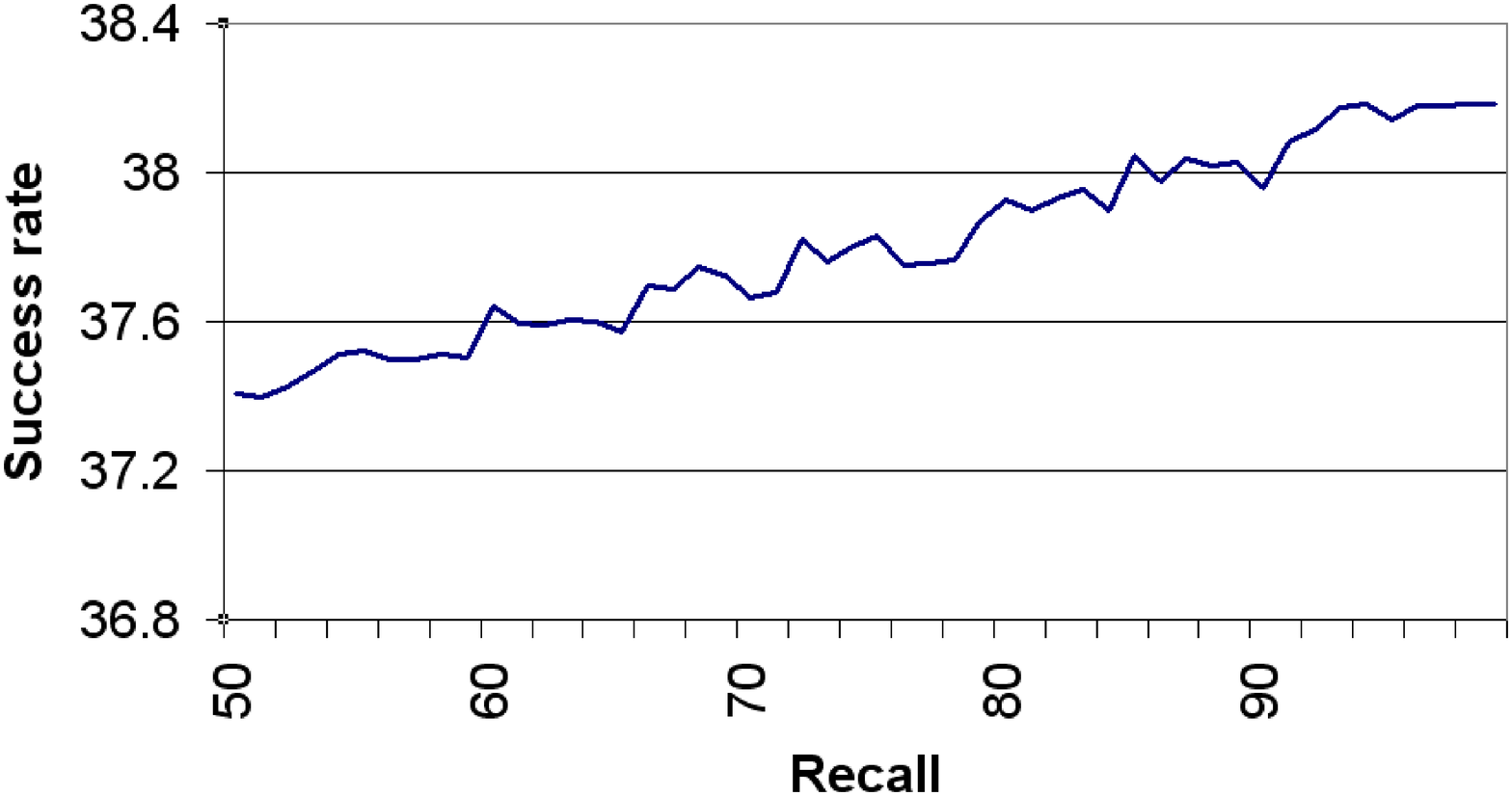}
\end{center}
\caption{The evolution of success rate with changes in precision and
recall} \label{fig:sim-data}
\end{figure*}

The results presented in the previous section makes it difficult to
have a clear idea about how accurate the animacy identifier needs to
be in order to have a significant positive influence on the
performance of {\sc mars}. In light of this, we performed an
experiment in which animacy identifiers which perform with a
predefined accuracy were simulated. These systems were designed in
such a way that the precision of animacy identification varies in
1\% increments from 10\% to 100\%, whilst recall varies from 50\% to
100\%.\footnote{We decided to control only the precision and recall
of animacy identification because in this way, indirectly, we also
control the recall and precision of the identification of inanimate
entities.} In order to achieve this, we introduced a controlled
number of errors in the annotated data by randomly changing the
animacy of a predetermined number of noun phrases. In order to
ensure fair results, the experiment was run 50 times for each
precision-recall pair, so that a different set of entities were
wrongly classified in each run. The list of classified entities (in
this case derived directly from the annotated data and not processed
by any of the methods described in this paper) was then used by {\sc
MARS} in the resolution process. Figure \ref{fig:sim-data} presents
the evolution of success rate as recall and precision are changed.
In order to see how the success rate is influenced by the increase in
recall, we calculated the success rates corresponding to the chosen
recall value and all the values for precision between 10\% and 100\%
and averaged them. In the same way we calculated the evolution of
success rate with changes in precision.

As can be seen in the figures, precision has a greater influence on
the success rate of {\sc mars} than recall because by increasing
precision, we notice an almost continuous increase in the
success rate. Overall, increasing recall also leads to an increase
in the success rate, but this increase is not smooth. On the basis
of this, we can conclude that high precision of animacy
identification is more important than recall. These results are also
supported by Table \ref{tab:extrinsic} where the Simple method leads
to good performance for {\sc mars} despite its low recall (but
higher precision) in the identification of animate entities.

\begin{figure*}[t]
\begin{center}
\includegraphics[height=11cm]{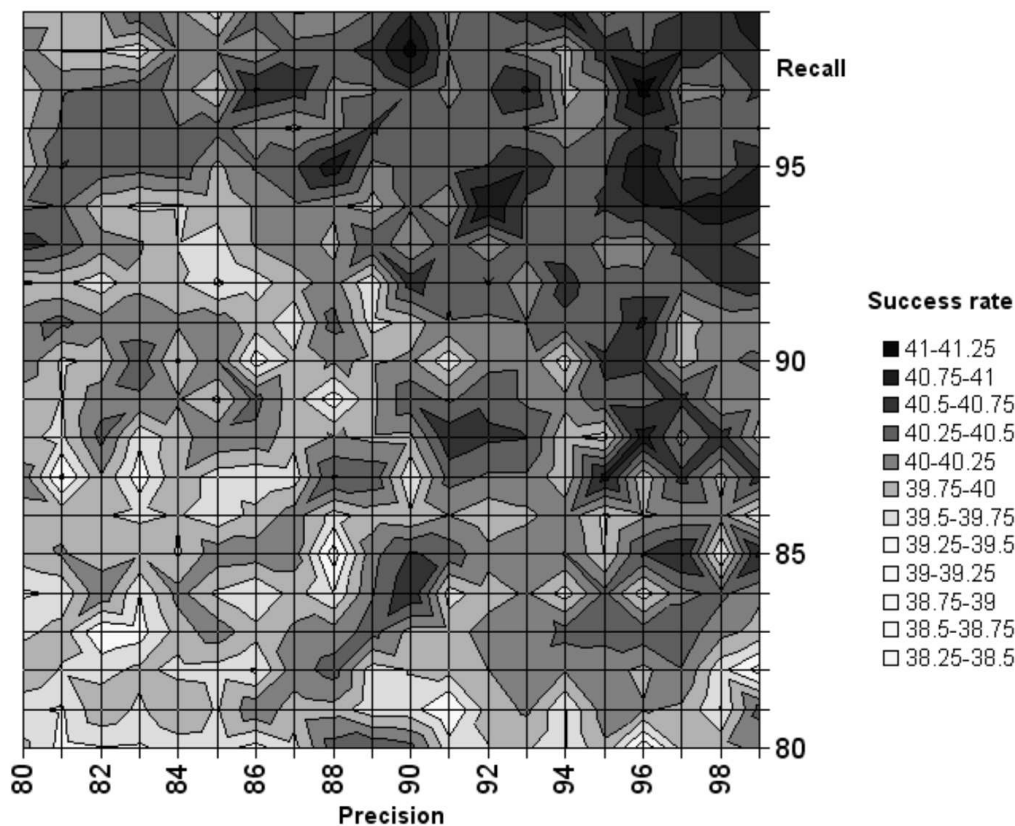}
\end{center}
\caption{Contour chart showing the success rate for different values
of precision and recall} \label{fig:sim-map}
\end{figure*}

Our experiments also reveal that for values higher than 80\% for
precision and recall, the success rate can vary considerably. For
this reason we decided to focus on this region. Figure
\ref{fig:sim-map} presents the success rate corresponding to
different precision-recall pairs using a contour chart. The darker
areas correspond to higher values of success rate. As noticed
before, the areas which correspond to high precision and high recall
also feature high success rates, but it is difficult to identify
clear thresholds for precision and recall which lead to improved
performance especially because most of the differences between the
first four intervals are not statistically significant.

\subsection{The Complexity of the Systems}\label{complexity}

One aspect which needs to be considered whenever a system is
developed is its complexity. This becomes a very important issue
whenever such a system is integrated with a larger system, which
needs to react promptly to its input (e.g. systems which are
available over the Web). In the present case, each method presented
in Section \ref{sec:methods} is more complex than the previous one,
and therefore requires more time to run. Table \ref{tab:times} shows
the time necessary to run each system on the two corpora. As can be
seen, the fastest method is the \emph{simple method} which has a
complexity proportional to \emph{n*m} where \emph{n} is the number
of entities in the entire corpus, and \emph{m} is the average number
of senses for each word in WordNet. The method which uses machine
learning is slower because it has to prepare the data for the
machine learning algorithm, a process which has a similar complexity
to the simple method, and in addition it has to run the memory-based
learning algorithm, which compares each new instance with all
instances already seen. Even though {\sc t}i{\sc mbl}, the machine
learning algorithm used, employs some sophisticated indexing
techniques to speed up the process, for large training sets, the
algorithm is slow. It has been noted that k-NN is an extremely slow
classifier and the use of alternate ML algorithms, such as maximum
entropy, may lead to quicker classification times with no loss in
accuracy. When word sense disambiguation is used, the processing
time increases dramatically, because the complexity of the algorithm
used is $n^m$ where $n$ is the number of distinct nouns from a text
to be disambiguated, and $m$ is the average number of senses from
WordNet for each noun. When the performance and run time of the
methods is considered, the best performing method is ML, which
ensures high accuracy together with relatively low execution time.
The use of an alternate WSD method that exploits N-best lists,
rather than considering all possible assignments of word senses,
would be likely to improve the speed of disambiguation. An approach
of this type has not yet been tested in our current work.

\begin{table}
\begin{center}
\begin{tabular}{|c|c|c|}
\hline
Method & AI & SEMCOR \\
\hline
Simple method & 3 sec. & 25 sec. \\
\hline ML & 51 sec. & 286 sec.\\
\hline Simple+WSD & \multicolumn{2}{c|}{Several hours}\\
\hline ML+WSD & \multicolumn{2}{c|}{Several hours}\\
\hline
\end{tabular}
\caption{The run time necessary for different
methods}\label{tab:times}
\end{center}
\end{table}

\section{Related Work}\label{sec:related}

With regard to work concerned with recognition of NP animacy, the
sole concern in this section is with those methods which tackle the
problem in English texts, a problem concerned with semantics that
cannot be addressed using morphological information, as it can be in
other languages.

Identification of the specific gender of proper names has been
attempted by Hale and Charniak \citeyear{hale-98}. That method works
by processing a 93931-word portion of the Penn-Treebank corpus with
a pronoun resolution system and then noting the frequencies with
which particular proper nouns are identified as the antecedents of
feminine or masculine pronouns. Their paper reports an accuracy of
68.15\% in assigning the correct gender to proper names.

The approach taken by Cardie and Wagstaff \citeyear{cardie-99} is
similar to the simple statistical one described in Section
\ref{sec:simple}, though the one described in this paper exploits a
larger number of unique beginners in the ontology, considers
semantic information about the verbs for which NPs are arguments,
and also considers all possible senses for each noun. In the
approach presented by Cardie and Wagstaff \citeyear{cardie-99},
nouns with a sense subsumed by particular nodes in the WordNet
ontology (namely the nodes \emph{human} and \emph{animal}) are
considered animate. In terms of gender agreement, gazetteers are
also used to assign each NP with a value for gender from the set of
\textsc{masc}uline, \textsc{fem}inine, \textsc{either} (which can be
assumed to correspond to animate), or \textsc{neuter}. The method
employed by
 Cardie and Wagstaff  is fairly simple and is incorporated as just one
feature in a vector used to classify coreference between NPs. The
employed machine learning method blindly exploits the value assigned
to the animacy feature, without interpreting it semantically.
WordNet has been used to identify NP animacy in work by Denber
\citeyear{denber-98}. Unfortunately, no evaluation of the task of
animacy identification was reported in those papers.

\section{Conclusions}\label{conclusions}

Animacy identification is a preprocessing step which can improve the
performance of anaphora resolution in English by filtering out
candidates which are not compatible, in terms of their agreement
features, with the referential expression. In this paper, a more
specific definition for animacy is used than the one proposed by
Quirk et al. \citeyear{quirk-85}. The adopted definition is more
appropriate and conveys the usefulness of this feature in anaphora
resolution. In the present study, the animacy of a noun phrase is
determined by the fact that it can be referred to by means of a
masculine or feminine pronoun as well as their possessive, relative
and reflexive forms.

In this paper, two different animacy identifiers were presented and
evaluated. The first one relies on the unique beginners from WordNet
in combination with simple rules to determine the animacy of a noun
phrase. Given that the unique beginners are too general to be used
in this way and that the rules were designed through na\"{i}ve
observations, a second method was proposed. This second approach
relies on a machine learning method and an enhanced WordNet to
determine the animacy of a noun phrase. In addition to the normal
semantic information, this enhanced WordNet contains information
about the animacy of a synset. This animacy information was
automatically calculated on the basis of manual annotation of the
SEMCOR corpus with animacy information.

The two animacy identifiers were evaluated using intrinsic and
extrinsic methods. The intrinsic evaluation employed several
measures to determine the most appropriate identifier. Comparison
between the results of these methods revealed that it is easy to
obtain relatively high overall accuracy at the expense of low
accuracy for the classification of animate references. For this
reason, it was concluded that the extra resources required by the
machine learning method, the best performing method, are fully
justified. Inter-annotator agreement was measured in order to
ascertain the difficulty of the task and as a result of this, it was
noted that the machine learning method reaches a level of
performance comparable to that of humans.

The extrinsic evaluation focused on how the performance of
{\sc mars}, a robust anaphora resolver, is influenced by the animacy
identifier. In light of the fact that {\sc mars} was designed to
resolve anaphors in texts from a different genre, the results
reported in the extrinsic evaluation did not focus only on the
accuracy of that system, but also on how many candidates are removed
by the animacy identifier and how many pronouns are left with no
valid antecedent to select from their sets of candidates as a result
of this process. Evaluation of {\sc mars} revealed that both of the
methods proposed in this paper improve its accuracy, but the degree
of improvement varies from one corpus to another. In terms of the
reduction of the number of candidates that the anaphora resolution
system has to consider, the machine learning method eliminates the
fewest candidates, but as a result it only evokes small increases in
the number of pronouns whose sets of competing candidates contain no
valid antecedents. For this reason, we argue that extrinsic
evaluation also shows that the machine learning approach is the most
appropriate method to determine the animacy of noun phrases.

Experiments with WSD produced mixed results. Only on one of the
corpora used in this research did it lead to small improvements in
performance. We thus conclude that the extra computation required in
order to disambiguate words is unnecessary.

\acks{We would like to thank Laura Hasler for helping us with the
annotation process and the three referees for their useful comments
which enabled us to improve the paper.}

\appendix 

\section{Tables}

\captionsetup{justification=centering}

\begin{table}[hb]
\centering
%\begin{center}
\mbox{
\begin{sideways}
\begin{tabular}{|c|c|c|c|c|c|c|c|}
\hline
 & \multicolumn{1}{c}{} & \multicolumn{3}{|c|}{Animacy} & \multicolumn{3}{c|}{Inanimacy} \\
\cline{3-8} \multicolumn{1}{|c|}{Experiment} &
\multicolumn{1}{c|}{Acc} & \multicolumn{1}{c|}{Prec} &
\multicolumn{1}{c|}{Recall} & F-meas & \multicolumn{1}{|c}{Prec} &
\multicolumn{1}{|c|}{Recall} & F-meas\\\hline\hline

\multicolumn{8}{|c|}{On AI corpus}\\\hline

Random baseline & \multicolumn{1}{c|}{50.60\%} &
\multicolumn{1}{c|}{19.37\%} & \multicolumn{1}{c|}{52.13\%} &
28.24\% & \multicolumn{1}{c|}{82.11\%} &
\multicolumn{1}{c|}{50.32\%} & 62.39\% \\\hline

Weighted baseline& \multicolumn{1}{c|}{31.01\%} &
\multicolumn{1}{c|}{18.07\%} & \multicolumn{1}{c|}{76.48\%} &
29.23\% &
\multicolumn{1}{c|}{79.27\%} & \multicolumn{1}{c|}{20.60\%} & 32.70\% \\
\hline

Dummy method& 82.77\% & 0\% & - & - & 82.77\% & 100\% & 90.57\%\\
\hline

Simple system & \multicolumn{1}{c|}{89.61\%} &
\multicolumn{1}{c|}{94.79\%} & \multicolumn{1}{c|}{52.69\%} &
67.73\% &
\multicolumn{1}{c|}{88.93\%} & \multicolumn{1}{c|}{99.24\%} & 93.80\% \\
\hline

Simple system + WSD & \multicolumn{1}{c|}{90.14\%} &
\multicolumn{1}{c|}{81.60\%} & \multicolumn{1}{c|}{62.57\%} &
70.83\% &
\multicolumn{1}{c|}{91.60\%} & \multicolumn{1}{c|}{96.66\%} & 94.06\% \\
\hline

Machine learning system & \multicolumn{1}{c|}{98.04\%} &
\multicolumn{1}{c|}{96.31\%} & \multicolumn{1}{c|}{92.19\%} &
94.20\% &
\multicolumn{1}{c|}{98.33\%} & \multicolumn{1}{c|}{99.26\%} & 98.79\% \\
\hline

Machine learning with WSD & \multicolumn{1}{c|}{97.85\%} &
\multicolumn{1}{c|}{95.37\%} & \multicolumn{1}{c|}{92.00\%} &
93.65\% &
\multicolumn{1}{c|}{98.34\%} & \multicolumn{1}{c|}{99.07\%} & 98.70\% \\
\hline\hline

\multicolumn{8}{|c|}{On SEMCOR corpus}\\\hline

Random baseline& \multicolumn{1}{|c|}{50.19\%} &
\multicolumn{1}{c|}{14.11\%} & \multicolumn{1}{c|}{50.49\%} &
22.05\% & \multicolumn{1}{c|}{86.19\%} &
\multicolumn{1}{c|}{50.14\%} & 63.39\%\\\hline

Weighted baseline & \multicolumn{1}{c|}{37.62\%} &
\multicolumn{1}{c|}{8.40\%} & \multicolumn{1}{c|}{74.44\%} & 15.09\%
&
\multicolumn{1}{c|}{88.41\%} & \multicolumn{1}{c|}{31.64\%} & 46.60\% \\
\hline

Dummy method  & 88.21\% & 0\% & - & - & 88.21\% & 100\% & 93.73\% \\
\hline

Simple system & \multicolumn{1}{c|}{91.42\%} &
\multicolumn{1}{c|}{88.48\%} & \multicolumn{1}{c|}{56.42\%} &
68.90\% &
\multicolumn{1}{c|}{91.81\%} & \multicolumn{1}{c|}{98.51\%} & 95.04\% \\
\hline

Simple system + WSD & \multicolumn{1}{c|}{93.33\%} &
\multicolumn{1}{c|}{88.88\%} & \multicolumn{1}{c|}{67.14\%} &
76.50\% &
\multicolumn{1}{c|}{93.94\%} & \multicolumn{1}{c|}{98.38\%} & 96.11\% \\
\hline

Machine learning system & \multicolumn{1}{c|}{97.72\%} &
\multicolumn{1}{c|}{91.91\%} & \multicolumn{1}{c|}{89.99\%} &
90.93\% &
\multicolumn{1}{c|}{98.75\%} & \multicolumn{1}{c|}{98.57\%} & 98.65\% \\
\hline

Machine learning with WSD  & \multicolumn{1}{c|}{97.51\%} &
\multicolumn{1}{c|}{89.97\%} & \multicolumn{1}{c|}{90.14\%} &
90.05\% &
\multicolumn{1}{c|}{98.59\%} & \multicolumn{1}{c|}{98.56\%} & 98.57\% \\
\hline

\end{tabular}
\end{sideways}
\hspace{1em}
\rotcaption{The results of the classification}\label{tab:intrinsic-res}
}

%\end{center}
\end{table}

\begin{table}
\begin{center}
\mbox{
\begin{sideways}
\begin{footnotesize}
\begin{tabular}{|c|c|c|c|}
\hline
System & Average candidates per pronouns & Percentage of pronouns without antecedent & MARS
accuracy\\\hline\hline

\multicolumn{4}{|c|}{Results on the AI Corpus: 215 animate pronouns} \\\hline
No filtering & 17.20 & 20.46 & 40.00\%\\\hline
Simple & 12.37 & 26.04 & 43.26\%\\\hline
Simple + WSD & 12.47 & 24.18 & 45.58\%\\\hline
Machine learning & 13.71 & 20.93 & 40.93\%\\\hline
Machine learning + WSD & 13.70 & 20.93 & 40.93\%\\\hline
Random baseline & 9.95 & 33.02 & 41.40\%\\\hline
Weighted baseline & 10.57& 40.46& 38.60\%\\\hline
Dummy method & 9.17 & 42.32 & 40.00\%\\\hline\hline

\multicolumn{4}{|c|}{Results on part of SEMCOR: 1250 animate pronouns} \\\hline
No filtering & 10.20 & 24.80 & 29.60\%\\\hline
Simple & 8.44 & 26.96 & 37.60\%\\\hline
Simple + WSD & 8.66 & 26.88 & 37.50\%\\\hline
Machine learning & 8.33 & 26.32 & 39.60\%\\\hline
Machine learning + WSD & 8.33 & 26.32 & 39.52\%\\\hline
Random baseline & 7.55 & 33.12 & 36.96\%\\\hline
Weighted baseline & 7.28 & 36.16 & 34.08\%\\\hline
Dummy method & 7.83 & 38.16 & 38.16\%\\\hline
\end{tabular}
\end{footnotesize}
\end{sideways}
\hspace{1em}
\rotcaption{The results of extrinsic evaluation}\label{tab:extrinsic}
}
\end{center}
\end{table}

% ------------------------------------------------------------------------
%GATHER{../../bibliography/files/bibliography}

\bibliographystyle{theapa}
\bibliography{../../../bibliography/files/bibliography}

\begin{thebibliography}{}

\bibitem[\protect\BCAY{Barbu, Evans, \BBA\ Mitkov}{Barbu
  et~al.}{2002}]{barbu-LREC-02}
Barbu, C., Evans, R., \BBA\ Mitkov, R. \BBOP2002\BBCP.
\newblock \BBOQ A corpus based analysis of morphological disagreement in
  anaphora resolution\BBCQ\
\newblock In {\Bem Proceedings of Third International Conference on Language
  Resources and Evaluation (LREC2002)}, \BPGS\ 1995 -- 1999\ Las Palmas de Gran
  Canaria, Spain.

\bibitem[\protect\BCAY{Barlow}{Barlow}{1998}]{barlow-DAARC-98}
Barlow, M. \BBOP1998\BBCP.
\newblock \BBOQ Feature mismatches and anaphora resolution\BBCQ\
\newblock In {\Bem Proceedings of DAARC2}, \BPGS\ 34 -- 41\ Lancaster, UK.

\bibitem[\protect\BCAY{Brennan, Friedman, \BBA\ Pollard}{Brennan
  et~al.}{1987}]{brennan-ACL-87}
Brennan, S.~E., Friedman, M.~W., \BBA\ Pollard, C.~J. \BBOP1987\BBCP.
\newblock \BBOQ A centering approach to pronouns\BBCQ\
\newblock In {\Bem Proceedings of the 25th Annual Metting of the ACL}, \BPGS\
  155 -- 162\ Stanford, California.

\bibitem[\protect\BCAY{Cardie \BBA\ Wagstaff}{Cardie \BBA\
  Wagstaff}{1999}]{cardie-99}
Cardie, C.\BBACOMMA\  \BBA\ Wagstaff, K. \BBOP1999\BBCP.
\newblock \BBOQ Noun phrase coreference as clustering\BBCQ\
\newblock In {\Bem Proceedings of the 1999 Joint SIGDAT conference on
  Emphirical Methods in NLP and Very Large Corpora (ACL'99)}, \BPGS\ 82 -- 89\
  University of Maryland, USA.

\bibitem[\protect\BCAY{Chomsky}{Chomsky}{1981}]{chomsky-81}
Chomsky, N. \BBOP1981\BBCP.
\newblock {\Bem Lectures on Government and Binding}.
\newblock Dordrecht: Foris.

\bibitem[\protect\BCAY{Cristea, Ide, Marcu, \BBA\ Tablan}{Cristea
  et~al.}{2000}]{cristea-COLING-00}
Cristea, D., Ide, N., Marcu, D., \BBA\ Tablan, V. \BBOP2000\BBCP.
\newblock \BBOQ An empirical investigation of the relation between discourse
  structure and co-reference\BBCQ\
\newblock In {\Bem Proceedings of the 18th International Conference on
  Computational Linguistics (COLING2000)}, \BPGS\ 208 -- 214\ Saarbrucken,
  Germany.

\bibitem[\protect\BCAY{Daelemans, Zavrel, van~der Sloot, \BBA\ van~den
  Bosch}{Daelemans et~al.}{2000}]{TiMBL-00}
Daelemans, W., Zavrel, J., van~der Sloot, K., \BBA\ van~den Bosch, A.
  \BBOP2000\BBCP.
\newblock \BBOQ {TiMBL}: Tilburg memory based learner, version 3.0, reference
  guide, ilk technical report 00-01\BBCQ\
\newblock Ilk\ 00-01, Tilburg University.

\bibitem[\protect\BCAY{Denber}{Denber}{1998}]{denber-98}
Denber, M. \BBOP1998\BBCP.
\newblock \BBOQ Automatic resolution of anaphora in {English}\BBCQ\
\newblock \BTR, Eastman Kodak Co, Imaging Science Division.

\bibitem[\protect\BCAY{Evans \BBA\ Or\u{a}san}{Evans \BBA\
  Or\u{a}san}{2000}]{evans-DAARC-00}
Evans, R.\BBACOMMA\  \BBA\ Or\u{a}san, C. \BBOP2000\BBCP.
\newblock \BBOQ Improving anaphora resolution by identifying animate entities
  in texts\BBCQ\
\newblock In {\Bem Proceedings of the Discourse Anaphora and Reference
  Resolution Conference (DAARC2000)}, \BPGS\ 154 -- 162\ Lancaster, UK.

\bibitem[\protect\BCAY{Fellbaum}{Fellbaum}{1998}]{fellbaum-98}
Fellbaum, C.\BED. \BBOP1998\BBCP.
\newblock {\Bem WordNet: An Electronic Lexical Database}.
\newblock The MIT Press.

\bibitem[\protect\BCAY{Frances \BBA\ Kucera}{Frances \BBA\
  Kucera}{1982}]{frances-82}
Frances, W.\BBACOMMA\  \BBA\ Kucera, H. \BBOP1982\BBCP.
\newblock {\Bem Frequency Analysis of {English} Usage}.
\newblock Houghton Mifflin, Boston.

\bibitem[\protect\BCAY{Hale \BBA\ Charniak}{Hale \BBA\
  Charniak}{1998}]{hale-98}
Hale, J.\BBACOMMA\  \BBA\ Charniak, E. \BBOP1998\BBCP.
\newblock \BBOQ Getting useful gender statistics from {English} text\BBCQ\
\newblock \BTR\ CS-98-06, Brown University.

\bibitem[\protect\BCAY{Hobbs}{Hobbs}{1976}]{hobbs-76}
Hobbs, J. \BBOP1976\BBCP.
\newblock \BBOQ Pronoun resolution\BBCQ\
\newblock Research report\ 76-1, City College, City University of New York.

\bibitem[\protect\BCAY{Hobbs}{Hobbs}{1978}]{hobbs-78}
Hobbs, J. \BBOP1978\BBCP.
\newblock \BBOQ Pronoun resolution\BBCQ\
\newblock {\Bem Lingua}, {\Bem 44}, 339--352.

\bibitem[\protect\BCAY{Kennedy \BBA\ Boguraev}{Kennedy \BBA\
  Boguraev}{1996}]{kennedy-COLING-96}
Kennedy, C.\BBACOMMA\  \BBA\ Boguraev, B. \BBOP1996\BBCP.
\newblock \BBOQ Anaphora for everyone: pronominal anaphora resolution without a
  parser\BBCQ\
\newblock In {\Bem Proceedings of the 16th International Conference on
  Computational Linguistics (COLING'96)}, \BPGS\ 113 -- 118\ Copenhagen,
  Denmark.

\bibitem[\protect\BCAY{Landes, Leacock, \BBA\ Tengi}{Landes
  et~al.}{1998}]{landes-98}
Landes, S., Leacock, C., \BBA\ Tengi, R.~I. \BBOP1998\BBCP.
\newblock \BBOQ Building semantic concordances\BBCQ\
\newblock In Fellbaum \cite{fellbaum-98}, \BPGS\ 199 -- 216.

\bibitem[\protect\BCAY{Lappin \BBA\ Leass}{Lappin \BBA\
  Leass}{1994}]{lappin-94}
Lappin, S.\BBACOMMA\  \BBA\ Leass, H.~J. \BBOP1994\BBCP.
\newblock \BBOQ An algorithm for pronominal anaphora resolution\BBCQ\
\newblock {\Bem Computational Linguistics}, {\Bem 20\/}(4), 535 -- 562.

\bibitem[\protect\BCAY{Mitchell}{Mitchell}{1997}]{mitchell-97}
Mitchell, T.~M. \BBOP1997\BBCP.
\newblock {\Bem Machine learning}.
\newblock McGraw-Hill Series in Computer Science. McGraw-Hill.

\bibitem[\protect\BCAY{Mitkov}{Mitkov}{1998}]{mitkov-98b}
Mitkov, R. \BBOP1998\BBCP.
\newblock \BBOQ Robust pronoun resolution with limited knowledge\BBCQ\
\newblock In {\Bem Proceedings of the 18th International Conference on
  Computational Linguistics (COLING'98/ACL'98)}, \BPGS\ 867 -- 875\ Montreal,
  Quebec, Canada.

\bibitem[\protect\BCAY{Mitkov}{Mitkov}{2002}]{mitkov-02}
Mitkov, R. \BBOP2002\BBCP.
\newblock {\Bem Anaphora resolution}.
\newblock Longman.

\bibitem[\protect\BCAY{Mitkov, Evans, \BBA\ Or\u{a}san}{Mitkov
  et~al.}{2002}]{mitkov-cicling-02}
Mitkov, R., Evans, R., \BBA\ Or\u{a}san, C. \BBOP2002\BBCP.
\newblock \BBOQ A new, fully automatic version of {Mitkov's} knowledge-poor
  pronoun resolution method\BBCQ\
\newblock In {\Bem Proceedings of CICLing-2002}, \BPGS\ 168 -- 186\ Mexico
  City, Mexico.

\bibitem[\protect\BCAY{Ng \BBA\ Cardie}{Ng \BBA\ Cardie}{2002}]{ng-02}
Ng, V.\BBACOMMA\  \BBA\ Cardie, C. \BBOP2002\BBCP.
\newblock \BBOQ Improving machine learning approaches to coreference
  resolution\BBCQ\
\newblock In {\Bem Proceedings of the 40th Annual Meeting of the Association
  for Computational Linguistics (ACL2002)}, \BPGS\ 104 -- 111\ Philadelphia,
  Pennsylvania.

\bibitem[\protect\BCAY{Quinlan}{Quinlan}{1993}]{quinlan-93}
Quinlan, J.~R. \BBOP1993\BBCP.
\newblock {\Bem C4.5: Programs for Machine Learning}.
\newblock Morgan Kaufmann.

\bibitem[\protect\BCAY{Quirk, Greenbaum, Leech, \BBA\ Svartvik}{Quirk
  et~al.}{1985}]{quirk-85}
Quirk, R., Greenbaum, S., Leech, G., \BBA\ Svartvik, J. \BBOP1985\BBCP.
\newblock {\Bem A Comprehensive Grammar of the English Language}.
\newblock Longman.

\bibitem[\protect\BCAY{Resnik}{Resnik}{1995}]{resnik-95}
Resnik, P. \BBOP1995\BBCP.
\newblock \BBOQ Disambiguating noun groupings with respect to {W}ordnet
  senses\BBCQ\
\newblock In Yarovsky, D.\BBACOMMA\  \BBA\ Church, K.\BEDS, {\Bem Proceedings
  of the Third Workshop on Very Large Corpora}, \BPGS\ 54--68\ Somerset, New
  Jersey. Association for Computational Linguistics.

\bibitem[\protect\BCAY{Sirkin}{Sirkin}{1995}]{sirkin-95}
Sirkin, R.~M. \BBOP1995\BBCP.
\newblock {\Bem Statistics for the social sciences}.
\newblock SAGE Publications.

\bibitem[\protect\BCAY{{Sparck Jones} \BBA\ Galliers}{{Sparck Jones} \BBA\
  Galliers}{1996}]{sparck-96}
{Sparck Jones}, K.\BBACOMMA\  \BBA\ Galliers, J.~R. \BBOP1996\BBCP.
\newblock {\Bem Evaluating natural language processing systems: an analysis and
  review}.
\newblock \BNUM\ 1083 in Lecture Notes in Artificial Intelligence. Springer.

\bibitem[\protect\BCAY{Tapanainen \BBA\ J{\"a}rvinen}{Tapanainen \BBA\
  J{\"a}rvinen}{1997}]{tapanainen-97}
Tapanainen, P.\BBACOMMA\  \BBA\ J{\"a}rvinen, T. \BBOP1997\BBCP.
\newblock \BBOQ A non-projective dependency parser\BBCQ\
\newblock In {\Bem Proceedings of the 5th Conference of Applied Natural
  Language Processing}, \BPGS\ 64 -- 71\ Washington D.C., USA.

\end{thebibliography}
%\bibliography{bib-complet,bib-new,summarization}

\end{document}